\documentclass[a4paper]{article}
\usepackage{siunitx}
\usepackage{INTERSPEECH2021}

\title{Reinforced Generative Adversarial Network for Abstractive Text Sum- marization}
\name{Tianyang Xu$^1$, Chunyun Zhang$^2$}
\address{
  $^1$Tianyang Xu\\
  $^2$Chunyun Zhang}
\email{xty.0532@gmail.com, zhangchunyun@sdufe.edu.cn}

\begin{document}

\maketitle
\begin{abstract}
  Sequence-to-sequence models provide a viable new approach to generative summarization, allowing models that are no longer limited to simply selecting and recombining sentences from the original text. However, these models have three drawbacks: their grasp of the details of the original text is often inaccurate, and the text generated by such models often has repetitions, while it is difficult to handle words that are beyond the word list. In this paper, we propose a new architecture that combines reinforcement learning and adversarial generative networks to enhance the sequence-to-sequence attention model. First, we use a hybrid pointer-generator network that copies words directly from the source text, contributing to accurate reproduction of information without sacrificing the ability of generators to generate new words. Second, we use both intra-temporal and intra-decoder attention to penalize summarized content and thus discourage repetition. We apply our model to our own proposed COVID-19 paper title summarization task and achieve close approximations to the current model on ROUEG, while bringing better readability.
\end{abstract}

\section{Introduction}

The process of using computers to process large amounts of text and generate concise and refined content is called text summarization. People can read the summary to grasp the main content of the original text, and because it reduces reading time, text summarization can bring considerable business value in time-sensitive work.
\\
\\
There are two main approaches to text summarization: extractive summarization and abstractive summarization. Extractive summarization approache writes summarizations based entirely on the original paragraphs (usually whole sentences) extracted directly from the source document, while the abstractive approache, which can mimic the way humans write summarization, has the potential to produce new words and phrases that are not in the source document. In general, the extractive approach is easier to implement because copying text directly from the source document ensures a baseline level of grammatically correctness and detail accuracy. However, the complex capabilities that are crucial for high-quality summaries, such as paraphrasing, generalizing, or incorporating real-world knowledge, can only be achieved in an abstractive summarization framework.
\\
\\
However, the vast majority of past work on text summarization has been extractive due to the considerable difficulty of abstractive summarization (Kupiec et al. 1995; Paice, 1990; Saggion and Poibeau, 2013). In recent years, the sequence-to-sequence model (seq2seq model, 2014) developed by Sutskever et al. has had success in solving the generative digest problem, where recurrent neural network algorithms (RNNs) can both read and freely generate text, which makes generative digests feasible (Chopra et al., 2016. Nallapati et al. 2016; Rush et al. 2015; Zeng et al. 2016). The above findings suggest that sequence-to-sequence models provide a new approach to generative summarization that can no longer be limited to simply selecting from the original text also being able to recombine sentences. However, in practice, it is found that such models are often less accurate in grasping the details of the original text, generating text with repetition, and still appear difficult when dealing with words beyond the vocabulary. Therefore, although these models are very promising, they still need to solve the problems of repetition, out-of-vocabulary (OOV) and detail errors.
\\
\\
Text summarization needs not only shorter sentences with similar semantic features, but we also need it to be precise enough to contain important details in the original text, i.e., the generated sentences are semantically consistent. To address these issues, this paper proposes an optimization model that incorporates a pointer-generator network approach, reinforcement learning, and adversarial generative network to better implement abstractive summarization.
\\
\\
For the task of summarizing long texts, the general sequence- to- sequence model often results in repeated words and inconsistent phrases. For this reason, we use an intra temporal attention mechanism in the encoder part of the generator and an intra decoder attention mechanism in the decoder part of the generator. In addition, when training sequence-to-sequence models with a minimized negative log-likelihood function, there is often the problem of exposure bias, i.e., training with supervised information about the next character, but testing without such supervised information. For this reason, we also combine maximum likelihood cross-entropy loss with the idea of reward from policy gradient reinforcement learning to alleviate this problem.
\\
\\
Similar to the work of Abigail et al. 2017, we introduce the approach of pointer-generator network in the decoder to retain more detailed information from the original text, such as match scores, or special numbers. The pointer-generator proposed by Bahdanau et al. is a combination of sequence-to-sequence model and pointer network ( The combination of sequence-to-sequence model and Pointer Network, on the one hand, maintains the ability of abstract generation through the sequence-to-sequence model, and on the other hand, improves the accuracy of abstracts and alleviates the problem of out-of-vocabulary by taking words directly from the original text through the pointer generator. The pointer generator is equivalent to dynamically adding words from the original text to the vocabulary in each abstract generation process, and the probability of selecting words occurring in the original text as abstracts is higher in each step of the prediction process compared to the simple sequence-to-sequence model.
\\
\\
In the text summarization task, most of the models introducing the reinforcement learning idea represented by Romain et al. refer to ROUGE, proposed by Chin-Yew Lin in 2004, as the reward function. Rouge (Recall-Oriented Understudy for Gisting Evaluation), a set of metrics for evaluating automatic abstracts and machine translation, is mainly based on recall, which is calculated by comparing an automatically generated abstract with a set of reference abstracts (usually manually written) and derives a corresponding score to measure the "similarity" between the automatically generated abstracts and the reference abstracts. ".
\\
\\
However, it is debatable whether ROUGE itself can be used as a measure of how well a model generates text. As Liu argues in his 2016 work, it is possible to improve the score of such discrete metrics without actually increasing readability or relevance. In an OpenAI's study, their model reached amazing level in manual evaluation, but its ROUGE value was not the highest. When the reference abstracts are not expertly designed and the number of reference abstracts available in the dataset is not big enough, linguistic diversity has a very strong impact on the results of metrics like ROUGE or BLEU and can seriously effect with the training of the model. For this reason, in Liu's work, the authors also recommend at least four reference samples for each participating sample in order to minimize the impact of linguistic diversity. However, in practice, such a requirement is difficult to achieve. Multiple reference samples require significant time and money costs to collect. Even in large and widely used datasets like CNN / Daily Mail, there is usually only one reference summary per document.
\\
\\
There is a framework that address this problem, Generative adversarial net (GAN). It is a framework proposed by (Goodfellow and others 2014). In GAN, a discriminative model D learns to distinguish whether a given data instance is real or not, and a generative model G learns to trick D by generating high quality data. This framework has been successful in computer vision tasks, such as style transfer and HD image generation, but not yet widely adopted in the field of natural language processing. In this paper, we refer to the design of GAN, and use a discriminator for the evaluation of the text generated by the model. The discriminator is modeled after the TextCNN proposed by Kim in 2014, and we also keep ROUGE as part of the reward in the training to ensure the consistency of the semantics, and control its weight in the total reward value by a predefined parameter. 
\\
\\
One of the major reasons why adversarial generative networks are not widely used in the field of natural language processing is that GAN is designed for generating continuous data such as images, but it does not do well on directly generating sequences of discrete tokens, such as texts. The reason behind this problem is that the generator need to sample a token from a probability distribution, and this operation is undifferentiable. This causes the gradient to fail to propagate backwards and thus the network weights cannot be updated. 
\\
\\
Another problem with GAN is that the discriminator can easily distinguish the unfinished sequences in the middle of training, which means the generator cannot know whether the previous step is good or not, and thus renders the scores from discriminator moot for the current training step. To solve this problem, our model refers to Sutton's policy gradient in 1999 and Yu et al.'s SeqGAN, which consider the sequence generation procedure as a sequential decision making process, and treats the generator as a policy in reinforcement learning and uses Monte Carlo tree search to complete multiple possible complete sentences and feed to the discriminator. The reward value of the generator for the current step is obtained by calculating the average of all the sentences scored. The weights of the generator are updated directly.
\\
\\
Experiments based on real data are conducted to investigate the readability and properties of the proposed model. In our experiment, the model outperforms or ties baseline models, with better readability concluded from human expert judgement.

\section{Our Model}

Our work fits the paradigm of a traditional adversarial generative network, the model design in this section will be developed in three directions: generator, discriminator, and the rollout modules. 

\subsection{Generator}

Generator is the most important component in a ganerative adversarial network model. It synthesis fake data and feed them to the discriminator to be distinguished. In this paper, our generator is an encoder-decoder structured sequence to sequence model. Encoder-decoder network was first introduced by Bengio et al. and Sutskever et al. in 2014 as an universal framework for sequence-to-sequence problems. What we did is build on the ground work of Yu et al., Abigail See et al, Paulus et al. and Nallapati et al., but with the following adaptations.

\subsubsection{Intra Temporal Attention}

Our baseline model is similar to that of Paulus et al. (2017). The encoder is a single-layer bidirectional LSTM and the decoder is a single-layer LSTM. The tokens of the article $w_i$ are fed into the encoder, producing a sequence of encoder hidden states $h_i$. For the i-th token in the input sequence $x=\{x_1, x_2, ...x_n\}$, this hidden state is defined as:

\begin{eqnarray}
h_{i}^{e}=\left[h_{i}^{e_{-} \mathrm{fwd}} \| h_{i}^{e_{-} \mathrm{bwd}}\right]
\end{eqnarray}
\\
where $||$ denotes concatenation operator of two vectors. On each decoding step $t$, we use an intra-temporal attention mechanism to attend over specific parts of the encoded input sequence in addition to the decoder’s own hidden state and the previously- generated word (Sankaran et al., 2016). While in training environment, the previous word is the previous word of the reference summary; while testing, it is the previous word output by the decoder. The decoder receives the word embedding of that previous word and has a decoder state $h_t^d$ . We define $e_{ti}$ as the attention score, which uses the dot-product between the two vectors, the attention distribution $\alpha_{t i}^{e}$ at decoding step $t$ and $i$-th word, can be calculated as:

\begin{eqnarray}
e_{t i}=\left\{\begin{array}{ll}
\exp \left(h_t^{d^T} * h_i^e\right) & \text { if } t=1 \\
\frac{\exp \left(h_t^{d^T} * h_i^e\right)}{\sum_{j=1}^{t-1} \exp \left(h_j^{d^T} * h_i^e\right)} & \text { if t > 1 }
\end{array}\right.
\end{eqnarray}

\begin{eqnarray}
\alpha_{t i}^{e}=\frac{e_{t i}}{\sum_{j=1}^{n} e_{t j}}
\end{eqnarray}
\\
The attention score $e _ {t i}$ penalizes input tokens that have obtained high attention scores in past decoding steps. This kind of attention prevents the model from attending over the same parts of the input sequence on different decoding steps. intra-temporal attention mechanism was previously shown in the work of Nallapati et al. (2016) and Paulus et al. (2017), and it was indicated that this type of mechanism can reduce repetitions when dealt with long documents. The attention distribution can be seen as a probability distribution over the source words. It is then used to produce a the context vector, which can be calculated:

\begin{eqnarray}
c_{t}^{e}=\sum_{i=1}^{n} a_{t i}^{e} h_{t i}
\end{eqnarray}

\subsubsection{Intra Decoder Attention}

Intra temporal attention mechanism allows different parts of the encoded input sequence to be used. However, there are still some repetation in the output sequence from time to time. This is because of the hidden state of the decoder. This problem is more frequently encontered when generating long sequences. To alleviate this problem, we include more information about the previously decoded sequence in the decoder based on the work of Paulus et al. (2017). This allows the model to make more sensible predictions while further panelizing the same information it already generated, thus prevent repetition.
\\
\\
For each decoding step t, the model computes a new decoder context vector $c^d_t$ . It can be calculated as:

\begin{eqnarray}
\alpha_{t t^{\prime}}^d =\left\{\begin{array}{ll}
\boldsymbol{0} & \text { if } t=1 \\
\frac{\exp \left(h_t^{d^T} * h_{t^{\prime}}^d\right)}{\sum_{j=1}^{t-1} \exp \left(h_t^{d^T} * h_{j}^d\right)} & \text { if t > 1 }
\end{array}\right.
\end{eqnarray}

\begin{eqnarray}
c_{t}^{d}=\sum_{j=1}^{t-1} \alpha_{t j}^{d} h_{j}^{d}
\end{eqnarray}

\subsubsection{Pointer-generator Network}

The main idea behind pointer-generator network is to use words from the original text to produce summaries that have less unknown tokens, therefor,  more accurate. In order to do so, the decoder need to be able to sample words from a distribution that contains both tokens from the original text as well as the ones produced by the encoder, which is a fixed vocabulary. At each decoding step $t$, the probibility distribution $P_{vocab}$ can be defined as:

\begin{eqnarray}
P_{vocab} = softmax( h^d_t || c^{e}_t || c^{d}_t )
\end{eqnarray}
\\
where, $h^d$ is the decoder state, $c^{e}$ and $c^{d}$ are context vectors produced by encoder and decoder. Context vectors can be seen as a representation of what has been read from the source for this step. The procuct is then fed through two linear layers to produce the vocabulary distribution $P_{vocab}$. 
\\
\\
In order to be able to copy words from source and deal with the problem of out-of-vocab, we introduced the idea of a pointer-generator (Abigail See et al; Gulcehre et al., 2016; Nallapati et al., 2016) , and use a soft switch $P_{gen}$ that decides at each decoding step whether to use the token generation or the pointer $P_{vocab}$.  At each decoding step $t$, we define $P_{gen} \in [0, 1]$ as:

\begin{eqnarray}
{P}_{gen} = \sigma (w_{c*}^Tc^*_t+w_s^Ts_t+w_x^Tx_t+b_{ptr})
\end{eqnarray}
\\
where $c_t^*$ is the context vector, $w_{c^*}$ , $w_s$, $w_x$ and $b_{ptr}$ are trainable parameters, and $\sigma$ is the sigmoid function. Context vector can be seen as a fixed sized semantic representation of the source text. The probability distribution of the source text can be seen as the attention distribution $a_t$, at decoding step $t$. This provides us with the final distribution from which the decoder would sample to predict words $w$:
\begin{eqnarray}
P(w)=p_{\text {gen }} P_{\text {vocab }}(w)+\left(1-P_{gen}\right) \sum_{i: w_{i}=w} a_{t i}
\end{eqnarray}
\\
If $w$ is an out-of-vocabulary word, then $P_{vocab}(w)$ is zero. Then the final distribution will contain only words from source text, preventing decoder from generating unknown token. This ability to produce OOV words is one of the primary advantages of pointer-generator models.
\\
\\
There are two mainstream training method for RNN: free-running mode and teacher-forcing mode. In the pretraining stage of the generator, we mainly use teacher-forcing. This is a quick and effective way to train RNN networks. It uses the ground truth data as the input for the next time step, instead of the output from last time step, thus making training to be faster. We define $y^{data} = \{y_1^{data}, y_2^{data}, . . . , y_n^{data} \}$ as the ground-truth output sequence for a given input sequence $x$. The maximum-likelihood training objective woule be:

\begin{eqnarray}
\min {-\sum_{t=1}^{n} \log p\left(y_{t}^{data} \mid y_{1}^{data}, \ldots, y_{t-1}^{data}, x\right)}
\end{eqnarray}

\subsection{Discriminator}

Another important component in the GAN framework is the discriminative model. In our model, the main objective of the discriminator is to distinguish the fake data generated by the generator from the given texts. 
\\
\\
In recent years, deep discriminative models such as deep neural network (DNN) (Vesely et al. 2013), convolutional neural network (CNN) (Kim 2014) and recurrent convolutional neural network (RCNN) (Lai et al. 2015) have shown a high performance in complicated sequence classification tasks. In this paper, we choose the TextCNN model proposed by Yoon Kim as our discriminator. TextCNN has recently been shown of great effectiveness in text (token sequence) classification (Zhang and LeCun 2015). Compared to traditional CNN networks for images, TextCNN has no significant change in the network structure. TextCNN contains a layer of convolution, a layer of max-pooling, and finally the output is connected to a softmax layer for binary classification. In this paper, we define $\mathbf{x}_{i} \in \mathbb{R}^{k}$ to be the k-dimensional word embedding of the i-th word in the input sequence. A padded sentence of $T$ can be represented as $\epsilon_{1:T} =x_1 \otimes x_2 \otimes ... \otimes x_T$, where $\bigoplus$ is the concatenation operator. Let $x_{1:1+j}$ to be the concatenation of words $x_i, x_{i+1}, ..., x_{i+j}$.  Then a convolutional kernel $ w\in \mathbb{R}^{h \times k}$, which is applied to a window of $h$ words to produce a new feature $c_i$, which can be defined as:

\begin{eqnarray}
c_i = \rho (\mathbf{w} \otimes \epsilon_{i:i+h-1} + b)
\end{eqnarray}
\\
where $b\in \mathbb{R}$ is a bias term, and $\rho$ is a non-linear function such as the hyperbolic tangent. We define $\otimes$ operator to be the summation of elementwise production. The feature map $c \in \mathbb{R}^{T-h+1}$ of the sentence $\mathbf{x}_{1:h}, \mathbf{x}_{1:h}, ... , \mathbf{x}_{T-h+1:T}$ can be represented as $ c = [c_1, c_2, ... , c_{T-h+1}] $, We then apply a max-over- time pooling operation (Collobert et al., 2011) over the feature map:

\begin{eqnarray}
\widetilde{c} = max\{ c_1, c_2,..., c_{T-h+1}\}
\end{eqnarray}
\\
To enhance the performance, we also add the highway architecture based on the pooled feature maps. Highway network was first poposed in 2015 in the work of Srivastava et al. (2015).  

\begin{eqnarray}
\begin{aligned}
\boldsymbol{\tau} &=\sigma\left(\boldsymbol{W}_{T} \cdot \tilde{\boldsymbol{c}}+\boldsymbol{b}_{T}\right), \\
\tilde{\boldsymbol{C}} &=\boldsymbol{\tau} \cdot F\left(\tilde{\boldsymbol{c}}, \boldsymbol{W}_{F}\right)+(\mathbf{1}-\boldsymbol{\tau}) \cdot \tilde{\boldsymbol{c}},
\end{aligned}
\end{eqnarray}
\\
where $W_T$ , $b_T$ and $W_F$ are highway layer weights, $F$ denotes an affine transform followed by ReLU and $\tau$ is the "transform gate" with the same dimensionality as $F (\tilde c, W_F )$ and $C$. Finally, we apply a sigmoid transformation to get the probability that a given sequence is real:
\begin{eqnarray}
\hat y = \sigma ( \mathbf{W}_o \cdot \tilde{\mathbf{C}} + b_o)
\end{eqnarray}
\\
where $W_o$ and $b_o$ is the output layer weight and bias, $\sigma$ is the sigmoid function. The optimization target is to minimize the cross entropy between the ground truth label and the predicted probability: 

\begin{eqnarray}
\mathbf{L}(y, \hat y) = - y \log \hat y - (1 - y) \log(1 - \hat y)
\end{eqnarray}
\\
where $y$ is the ground truth label of the input sequence and $\hat y$ is the predicted probability from the discriminative models.

\subsection{Rollout Module}

Most discriminative models can only perform classification well for an entire sequence rather than the unfinished one. Therefore, following the work of Yu et al. (2017), we added a rollout module to the generative model to complete the unfinished sequence in intermediate steps. This module simulates the rest of the sequence multiple times to get multiple complete sentence. Then the sentences are fed into the discriminator to get each of the classification result. The overall reward is then obtained by calculating the average of all the results. 
\\
\\
The rollout module used in this paper is an exact copy of the generator, but in the experiment environment, it uses a copy of the generator that is updated with a several step lag, thus increasing the stability of the long sequence generation in reinforcement learning.  The rollout module is not directly updated via gradient descent, but rather updates its parameters manually.

\subsection{Reinforcement Learning and Policy Gradient}

According to the approach introduced in SeqGAN by Yu et al. 2017 and Sutton et al, 1999, the optimization objective of the generator model (i.e., the policy in the reinforcement learning) is $G_\theta (y_t |Y_{1:t-1} )$, i.e., the expectation of generating a reward for a certain complete text sequence conditional under the conditions of $s_0$ and $\theta$, the expected reward for producing a certain complete text sequence:

\begin{eqnarray}
\mathbb{E}[R_{T} \mid s_{0}, \theta]=\sum_{y_{1} \in \mathcal{Y}} G_{\theta}\left(y_{1} \mid s_{0}\right) \cdot R_{D_{\phi}}^{G_{\theta}}\left(s_{0}, y_{1}\right)
\end{eqnarray}
\\
\\
where $R_{T}$ is the reward for the generated complete sequence. This reward value is derived from the discriminator $D_\phi$. $R_{D_\phi}^{G_\theta}(s, a)$ is the action-value function of a tragectory, i.e., the expectated reward obtained by the $G_\theta$ policy generating act $a$, starting from state $s$. And starting from a given initial state, the objective of the generator is to generate a sequence that the discriminator will consider to be true. We set the probability of a sample being judged true by the discriminator $D_\phi(Y_{1:T}^n)$ as the reward value. That is:

\begin{eqnarray}
R_{D_\phi}^{G_\theta}(a = y_T, s = Y_{1:T-1})=D_\phi(Y_{1:T})
\end{eqnarray}
\\
However, since the discriminator can only deal with complete sequences, the reward value is only available when $y_1$ reaches $y_{T-1}$. This is because what we need is a long-term reward (i.e., a reward for a whole sentence), which is similar to how in a chess game each move has an effect on the outcome, but we only care about the final result. Therefore, to calculate the value of each action, we use a Monte Carlo tree search algorithm and a completing strategy (i.e., the rollout module in the generative model), $G_\beta$, to generate the $T - t$ characters after the current action $t$ up to the last action $T$ . As this is a Monte Carlo search, each time multiple outputs are possible, we calculate the reward values for all possible sequences that are completed using Monte Carlo search after the same $y_1$ and average them as follows:

\begin{eqnarray}
\begin{array}{ll}
R_{D_{\phi}}^{G_{\theta}}\left(s=Y_{1: t-1}, a=y_{t}\right)=\\ 
\left\{\begin{array}{ll}
\frac{1}{N} \sum_{n=1}^{N} D_{\phi}\left(Y_{1: T}^{n}\right), Y_{1: T}^{n} \in \mathrm{MC}^{G_{\beta}}\left(Y_{1: t} ; N\right) & \text { for } t<T \\
D_{\phi}\left(Y_{1: t}\right) & \text { for } t=T
\end{array}\right.
\end{array}
\end{eqnarray}
\\
where $Y_{1:t}^n = (y_1, ... y_n)$ and $y_{t+1:T}^n$ are generated based on the rollout policy $G_\beta$ and the current state. In our experiments, $G_\beta$ has the same network structure and similar weights as the generator. $n$ is the number of complement operations, i.e. the rollout module runs $n$ times from the current state to the end of the generated sequence, providing $n$ complete sequences.
\\
The discriminator $D_\phi$ is trained as:

\begin{eqnarray}
\min _{\phi}-\mathbb{E}_{Y \sim p_{\text {data }}}\left[\log D_{\phi}(Y)\right]-\mathbb{E}_{Y \sim G_{\theta}}\left[\log \left(1-D_{\phi}(Y)\right)\right]
\end{eqnarray}
\\
\\
The discriminator is trained on both generated data and real data, therefore it can be noted that the training of such a discriminator will be limited by the training of the generator, i.e., it requires the generator to generate a sufficient number of samples that resemble the true summary. So in the experiment, the generator model needs to be trained in a supervised manner before we can train the discriminator and perform adversarial training. After one or more rounds of training of the discriminator $D_\phi$, a better discriminator $D_\phi$ is obtained, at which point the discriminator $D_\phi$ is used to update the generator $G_\theta$. 
\\
\\
And the update of $G_\theta$ can use gradient descent. The objective of a certain agent, such as $G$, in a reinforcement learning algorithm is to maximize the expected reward. Given a starting state $h_i$, let's assume the parameter of the agent is $\theta$, and the output of the agent is $x_i$. It interacts with the environment (or in this case, the discriminator) and generate a reward of $R_i(h_i,x_i)$. The expected reward of the whole action-state trajectory can be described as:

\begin{eqnarray}
\overline{R_\theta} = \sum_{h} P(h) \sum_{x} R(h,x) P_{\theta}(x\mid h)
\end{eqnarray}
\\
\\
where $P(h)$ means the possibility of the starting state $h$. And the objective of this model would be:

\begin{eqnarray}
\theta^* = \mathop{argmax}\limits_{\theta}\overline{R_\theta}
\end{eqnarray}
\\
\\
The expected reward $\overline{R_\theta}$ can be calculated as:

\begin{eqnarray}
\overline {R_{\theta}}&=&\sum_{h} P(h) \sum_{x} R(h,x) P_{\theta}(x\mid h)  \nonumber    \\
~&=& E_{h\sim P(h)}[E_{x\sim P_{\theta}(x \mid h)}R(h,x)] \nonumber    \\
~&=& E_{h\sim P(h), x\sim P_{\theta}(x\mid h)}R(h,x)
\end{eqnarray}
\\
\\
An RNN based generator may not give the same output given multiple inputs of the same stating state. Meanwhile, it is also impossible to sample every possible imput state. So the $h$ and $x$ cannot be integrated, which renders the function above moot. But given the same input state, the possibility distribution produced by the generator remains the same, and the input states can be sampled multiple times to approximate the actual possibility distribution of $h$. Let each sample of $h$ and $x$ be $(h^i,x^i)$, and the function above can be approximated as:

\begin{eqnarray}
\overline {R_{\theta}} \approx \frac{1}{N}\sum_{i=1}^{N}R(h^i,x^i)
\end{eqnarray}\\
\\
In order to utilize gradient ascent to maximize [eq objective], we need to differentiate $\overline{R_{\theta}}$ by $\theta$. However, after approximation, $\overline{R_{\theta}}$ is independent of $\theta$. This is because the $\theta$ is implicitly included in the sample distribution. However we can circumvent this limitation by taking the idea of Policy Gradient.

\begin{eqnarray}
\nabla \overline {R_{\theta}}&=&\sum_{h} P(h) \sum_{x} R(h,x) \nabla P_{\theta}(x\mid h)  \nonumber    \\
~&=& \sum_{h} P(h) \sum_{x} R(h,x) P_{\theta} (x\mid h) \frac{\nabla P_{\theta}(x\mid h)}{P_{\theta}(x\mid h)} \nonumber    \\
~&=& \sum_{h} P(h) \sum_{x} R(h,x) P_{\theta} (x\mid h) P_{\theta}(x\mid h)  \nonumber\\
~& & \nabla\log_{}{P_{\theta}(x \mid h)}\nonumber    \\
~&=& E_{h\sim P(h), x\sim P_{\theta}(x\mid h)} [ R(h,x)\nabla\log_{}{P_{\theta}(x \mid h)} ]\nonumber    \\
~&\approx& \frac{1}{N}\sum_{i=1}^{N}R(h^i,x^i) \nabla \log_{}{P_{\theta}(x^i \mid h^i)}\nonumber
\end{eqnarray}
\\
In our model, the reward $R$ is assigned by the discriminator $D_{\phi}$. Therefore:

\begin{eqnarray}
\nabla \overline {R_{\theta}}&\approx& \frac{1}{N}\sum_{i=1}^{N}R(h^i,x^i) \nabla \log_{}{P_{\theta}(x^i \mid h^i)}\nonumber    \\
~ &\approx& \frac{1}{N}\sum_{i=1}^{N}(D_{\phi}(h^i,x^i)-b) \nabla \log_{}{P_{\theta}(x^i \mid h^i)}\nonumber    \\
\end{eqnarray} \\
where $b$ stands for a baseline reward. For a sequence like a sentence with multiple steps (words), it is usually difficult to determine the reward for every step. To achieve word-level reward value, we can further describe the reward function for each word as:

\begin{eqnarray}
\nabla \overline {R_{\theta}}&\approx& \frac{1}{N}\sum_{i=1}^{N}R(h^i,x^i) \nabla \log_{}{P_{\theta}(x^i \mid h^i)}\nonumber    \\
~ &\approx& \frac{1}{N}\sum_{i=1}^{N}\sum_{i=1}^{T}(Q(h^i,x^i_{1:t})-b) \nabla \log_{}{P_{\theta}(x^i \mid h^i_{1:t})}\nonumber    \\
\end{eqnarray} \\
\\
where $T$ is the length of the sequence and $t$ is the current step. $Q(h^i,x^i_{1:t})$ can be estimated through Monte Carlo search. \\
\\
Then we can update the model $\theta$ using this function:\\

\begin{eqnarray}
{\theta}^{new} = \theta^{old} + \eta_{h}\nabla\overline {R_{{\theta}^{old}}}
\end{eqnarray}
\\
where $\eta_h \in \mathbb{R}^+$ denotes the learning rate at $h\text{-th}$ step. \\

\section{Related work}

\subsection{Neural Networks for Text Summarization}
Since the first modern neural network approach was introduced into abstractive text summarization in the work of Rush et al. (2015), the field has grown significantly. Rush et al.'s work (2015) achieved optimal performance on two sentence-level summarization datasets, DUC-2004 and Gigaword. Their approach, with attention mechanisms at its core, further improves the performance of these datasets through abstract meaning representations (Takase et al., 2016), recurrent decoders (Chopra et al., 2016), hierarchical networks (Nallapati et al., 2016), variable autoencoders (Miao and Blunsom, 2016), and direct optimization of performance metrics (Ranzato et al. ), have further improved the performance of these datasets. Nallapati et al. (2016) adapted the DeepMind Q\&A dataset (Hermann et al., 2015 for summarisation to form the CNN/Daily Mail dataset and provided the first baseline score for abstractive summarization. Subsequently, the same authors published a neural extraction method (Nallapati et al., 2017) which used hierarchical RNNs to select sentences and found that it significantly outperformed their abstraction results in terms of the ROUGE metric.

\subsection{Pointer Network}
A pointer network (Vinyals et al., 2015) is a sequence-to-sequence model that uses the soft attention distribution of Bahdanau et al. (2015) to produce an output sequence consisting of elements of the input sequence. Pointer networks have been used to create hybrid approaches to NMT (Gulcehre et al., 2016), language modelling (Merity et al., 2016) and summarisation (Gu et al., 2016; Gulcehre et al., 2016; Miao and Blunsom, 2016; Nallapati et al., 2016; Zeng et al., 2016).

\subsection{Reinforcement Learning}
This is a way of training a model (agent) to interact with a given environment in order to maximise the reward. Reinforcement learning has been used to solve a wide variety of problems, typically when the model has to perform discrete actions before it can obtain a reward, or when the metric to be optimised is unguidable and traditional supervised learning methods cannot be used. And this is well suited for sequence generation tasks, as many of the metrics used to evaluate these tasks (such as BLEU, ROUGE or METEOR) are not differentiable.
\\
\\
In order to optimise models directly using these metrics, Ranzato et al. (2015) applied a reinforcement algorithm (Williams, 1992) to train various RNN-based models for sequence generation tasks, which brought significant improvements compared to previous supervised learning methods. While their approach requires an additional neural network, called a discriminator model, to predict reward expectation and objective function gradients, Rennie et al. (2016) devised a self-discriminating sequence training method that does not require this discriminator model, and further improved it in an application to the image captioning task.

\section{Dataset}

We used the CNN/Daily Mail dataset (Hermann et al., 2015; Nallapati et al., 2016), which consists mainly of news texts extracted from CNN and Daily Mail (781 words each on average), and their respective abstracts (average 3.75 words per abstract, approximately 56 words). We used the preprocessing script provided by Nallapati et al. (2016) to obtain similar data, resulting in a training set of 287,226 paired samples, a validation set of 13,368 paired samples, and a test set of 11,409 samples.
\\
\\
Also, to ensure that the same model is usable on low word frequency, specialised domain texts, we constructed a COVID-19 dataset consisting of paper abstracts and paper titles based on the COVID-19 paper database published by Kaggle. On 7 December 2020, we extracted additional data from the updated database and pre-processed it to obtain 38540 paired samples for the test and validation sets. The test and validation sets were.

\section{Experiment}
\subsection{Experiment Procedure}
There are four main procedures of training this model: pre-training generator, pre-training discriminator, adversarial training and discriminator update.
\\
\\
In generator pre-training, we use the Maximum Likelihood Estimation method to pre-train the generator G. After pre-training, the generator and the discriminator are trained separately. Then we train generator in an adversarial manner. After certain steps of training and updating the generator. The discriminator needs to be re-trained separately to adapt to the updated generator. The discriminator is trained with positive samples from the dataset and negative samples from the generator. In order to keep the balance, the number of negative examples we generate for training discriminator is the same as the positive examples. Further more, to reduce the variability of the estimation, we use different sets of negative samples combined with positive ones, which is similar to bootstrapping (Quinlan 1996).

\subsection{Experiment Setup}

For all experiments, we restricted the vocabulary size to 50,000 words, small in line with Nallapati et al. 's (2016) 150,000 words/60,000 words, similar to Pointer-Network. As we trained primarily on our COVID-19 dataset, we limited the input and output lengths to 15 and 55 words, respectively. We used a 256-dimensional word2vec word embedding as input and fixed the hidden state of the model to 512 dimensions.
\\
\\
Similar to the model training in Nallapati et al. (2016), we used pre-trained word vectors, which differs from the baseline model (Pointer Network) which is not pre-trained. We trained using the Adam optimizer, rather than the AdaGrad used by Abigail See, which was proposed by Kingma and Lei Ba in December 2014. It combines the advantages of both AdaGrad and RMSProp optimisation algorithms. The Adam optimiser has a better performance for models with unstable training such as adversarial generative networks. The Adam optimizer works better for models with unstable training like the counteracting generative networks. Also, we set the gradient clipping to prevent gradient explosion, set the maximum number of normals to 2 and do not use regularisation.
\\
\\
For training and testing on CNN/DM dataset, we truncated the article length to 200 words and limited the summary length to 25 tokens for training and 25 tokens for testing, in order to speed up training and testing. We trained on cloud computing platforms provided by Google Colab and others, on a single, batch size of 200, with a predominantly Tesla T4 GPU.
\\
\\
When training the generators, we trained the model for approximately 600,000 iterations (30 rounds), which is similar to the 35 rounds required for the best model by Nallapati et al. (2016) and approximates the 33 rounds required for Abigail See's model. The training time for the vocabulary model was around 3 days and 18 hours when the vocabulary size was set at 50,000. In adversarial training, due to the more complex network model, the data generated at each time was much higher than the generator pre-training, which took 4 rounds in approximately the same amount of time (3 days and 12 hours).

\begin{table}[]
\caption{Results from COVID-19 Paper Title Dataset}
\label{tab:my-table}
\begin{tabular}{@{}ll@{}}
\toprule
{{\begin{tabular}[c]{@{}l@{}}Original\\Document\end{tabular}}} & {\begin{tabular}[c]{@{}l@{}}
the recent global pandemic of covid-19 is \\increasingly alarming. as of June 21, there are \\more than 8.7 million worldwide cases with \\460000 deaths. Nepal is not an exception to \\covid-19 and is currently facing a challenge to \\prevent the spread of infection. the analysis of \\the detected cases, severity, and outcomes of the \\cases within a country is important to have a \\clear picture of where the pandemic is heading \\and what measures should be taken to curb the \\infection before it becomes uncontrollable. \\in this manuscript, we have covered all the cases\\, recoveries, and deaths attributed to covid-19 \\in Nepal starting from the first case on January \\23 to June 21 . at present, covid-19 has spread all\\over Nepal with a rapid increase in the number \\of new cases and deaths which is very alarming \\in a low - income country with an inadequate \\health care system like Nepal. although the \\government implemented an early school closure\\ and lock - down, the management to contain \\covid-19 does not appear to be adequate. our \\manuscript gives a clear understanding of the \\current situation of covid-19 in Nepal which is \\important for providing a direction towards \\proper management of the disease.
\end{tabular}}\\ \midrule
{Reference}                  & {covid-19: the current situation in Nepal}  \\
{Our model}                  & {{\begin{tabular}[c]{@{}l@{}}cases, recoveries, and deaths by covid-19 in \\Nepal and proper management of the disease\end{tabular}}}  \\
{Pointer-network}            & {{\begin{tabular}[c]{@{}l@{}}manuscript of covid-19 cases, recoveries, and \\deaths in Nepal\end{tabular}}} 
\end{tabular}
\end{table}

\section{Result}

Our results are presented in Tables 1, 2 and 3. We evaluated our model using the standard ROUGE metric (Lin, 2004b), reporting F1 scores for ROUGE-1, ROUGE-2 and ROUGE-L (measuring word overlap, large word overlap and longest common sequence between the reference summary and the summary to be evaluated, respectively). We used the pyrouge software package to calculate the ROUGE scores.
\\
\\
Given that we generated plain text summaries but Nallapati et al. (2016; 2017) generated anonymised summaries (replacing all entities), our ROUGE scores are not strictly comparable. There is evidence that the original dataset may have resulted in generally higher ROUGE scores than the anonymised dataset. One possible explanation is that multiple word-named entities may lead to higher n-gram overlap rates. Unfortunately, ROUGE is the only available means of comparison with the work of Nallapati et al. 
\\
\\
Also, due to the reduced or eliminated share of ROUGE score in the reward function for reinforcement learning training, it can be seen that our proposed model sometimes scores inferior to the pointer generator used by Abigail See in terms of ROUGE scores. However, such scores do not necessarily mean that the quality of the text generated is inferior to that of the latter. As we have discussed before, the ROUGE function does not directly equal the quality of the generated text, but rather its difference to the reference text. Since our COVID-19 dataset generates paper titles from paper abstracts, the generated texts are generally short single-sentence texts, which are less difficult than the CNN/DM dataset and will score a little higher in ROUGE.

\begin{table}[]
\caption{Results on CNN/Daily Mail Dataset}
\label{tab:example}
\centering
\begin{tabular}{llll}
\toprule
Models & {\textbf{ROUGE-1}} & {\textbf{ROUGE-2}} & {\textbf{ROUGE-L}} \\
\midrule
{{{\begin{tabular}[c]{@{}l@{}}Romain Paulus\\(ML+RL)\end{tabular}}}}  & {39.87}            & {15.82}            & {36.90}            \\
{Pointer-Generator}      & {39.53}            & {17.28}            & {36.38}            \\
{{\begin{tabular}[c]{@{}l@{}}Our model\\(50K, No RL)\end{tabular}}} & {39.45}            & {16.02}            & {36.33}            \\
{{\begin{tabular}[c]{@{}l@{}}Our model\\(50K, RL)\end{tabular}}}    & {37.89}            & {15.67}            & {34.99}            \\ \bottomrule
\end{tabular}
\end{table}

\begin{table}[]
\caption{Results from COVID-19 Paper Title Dataset}
\label{tab:my-table}
\centering
\begin{tabular}{@{}ll@{}}
\toprule
{\textbf{Models}}        & {\textbf{ROUGE-1}} \\ \midrule
{Romain Paulus (ML+RL)}  & {57.87}            \\
{Pointer-Generator}      & {60.81}            \\
{Our model (50K, No RL)} & {57.99}            \\
{Our model (50K, RL)}    & {54.45}            \\ \bottomrule
\end{tabular}
\end{table}


\section{Discussion}

Although our model achieves the expected results, it still has shortcomings. After extensive testing and validation, the following conclusions were obtained.
\\
\\
1) Even if a more complex network is used instead of the original ROUGE, the reward function design is still relatively simple. The effectiveness of using a complex TextCNN for text classification tasks does not mean that it measures up to the standard of human judgments on text summarization. Even with human observation, it is often difficult to determine which of the two texts is more appropriate. Therefore to achieve the most scientific means of discrimination, ideally multiple sets of manual scoring should be used instead of the existing reward function.
\\
\\
2) Secondly, based on our observations, with the addition of these new techniques, despite reducing the original problems of repetition and inaccuracy, the generated text makes extensive use of the original content, resulting in it being more akin to an extractive rather than a abstractive summary. We believe that such a result may be due to the size and natural limitations of word vectors to characterise more complex and diverse word groups. We therefore believe that further improvements can be made to the word vectors, such as the use of dynamic word vectors, to achieve more desirable results.
\\
\\
3) Neither reinforcement learning nor adversarial generative networks, have a direct improvement on the generative model, and their impact is mostly on fine-tuning the network at the end of training. Therefore, proposing a more efficient generator model may lead to more substantial improvements.

\section{Conclusion}

In this work, we apply attentional encoder-decoders, generative adversarial networks and reinforcement learning to the task of abstractive text summarization with satisfactory results. The methods we apply all solve or somewhat alleviate a specific problem in generative summarization, thus further improving readability. We also presented a new multi-sentence dataset and established benchmark figures. As part of future work, we plan to focus our efforts on building more robust models for multi-sentence composition summaries.



\section{Acknowledgements}


This work is supported by the National Natural Science Foundation of China (61703234, 61701281, 61876098), Shandong Provincial Natural Science Foundation Key Project (ZR2020KF015), Major Scientific and Technological Innovation Projects of Shandong Province (2018CXGC1501) and the Fostering Project of Dominant Discipline and Talent Team of Shandong Province Higher Education Institutions.



\end{document}